\documentclass[conference]{IEEEtran}
\IEEEoverridecommandlockouts
\usepackage{cite}
\usepackage{amsmath,amssymb,amsfonts}
\usepackage{algorithmic}
\usepackage{graphicx}
\usepackage{textcomp}
\usepackage{caption}
\usepackage{booktabs}
\usepackage[table]{xcolor} 
\usepackage[most]{tcolorbox} 
\usepackage{placeins} 
\usepackage{bm}
\usepackage{hyperref}

\def\BibTeX{{\rm B\kern-.05em{\sc i\kern-.025em b}\kern-.08em
    T\kern-.1667em\lower.7ex\hbox{E}\kern-.125emX}}
\begin{document}

\title{Bailicai: A Domain-Optimized Retrieval-Augmented Generation Framework for Medical Applications\\
\thanks{ Yongbin  is corresponding author.}
}

\author{
	\IEEEauthorblockN{Cui Long}
	\IEEEauthorblockA{\textit{University of South China}\\
		cl1716649290@stu.usc.edu.cn}
	\and
	\IEEEauthorblockN{Yongbin Liu}
	\IEEEauthorblockA{\textit{University of South China}\\
		yongbinliu03@gmail.com}
	\and
	\IEEEauthorblockN{Chunping Ouyang}
	\IEEEauthorblockA{\textit{University of South China}}
	\and
	\IEEEauthorblockN{Ying Yu}
	\IEEEauthorblockA{\textit{University of South China}}
}

\maketitle

\begin{abstract}
Large Language Models (LLMs) have exhibited remarkable proficiency in natural language understanding, prompting extensive exploration of their potential applications across diverse domains. In the medical domain, open-source LLMs have demonstrated moderate efficacy following domain-specific fine-tuning; however, they remain substantially inferior to proprietary models such as GPT-4 and GPT-3.5. These open-source models encounter limitations in the comprehensiveness of domain-specific knowledge and exhibit a propensity for 'hallucinations' during text generation. To mitigate these issues, researchers have implemented the Retrieval-Augmented Generation (RAG) approach, which augments LLMs with background information from external knowledge bases while preserving the model's internal parameters. However, document noise can adversely affect performance, and the application of RAG in the medical field remains in its nascent stages. This study presents the Bailicai framework—a novel integration of retrieval-augmented generation with large language models optimized for the medical domain. The Bailicai framework augments the performance of LLMs in medicine through the implementation of four sub-modules. Experimental results demonstrate that the Bailicai approach surpasses existing medical domain LLMs across multiple medical benchmarks and exceeds the performance of GPT-3.5. Furthermore, the Bailicai method effectively attenuates the prevalent issue of hallucinations in medical applications of LLMs and ameliorates the noise-related challenges associated with traditional RAG techniques when processing irrelevant or pseudo-relevant documents.
\end{abstract}


\section{Introduction}
Large language models (LLMs), trained on extensive datasets, demonstrate the ability to generate plausible responses through their internal parameterized knowledge, achieving performance comparable to that of human experts in complex natural language processing tasks and specialized domains\cite{llmsurvey,llmfl,openai,palm}. Recent studies, including those on GPT-4\cite{openai,medprompt} and Med-PaLM 2\cite{medpalm2}, indicate that these proprietary LLMs can achieve expert-level performance on specific medical benchmarks. The sensitive nature of medical data, however, raises concerns regarding patient privacy risks associated with these closed-source models\cite{llmmedicine}, thereby prompting heightened interest in the application of open-source LLMs in the medical domain. Nevertheless, open-source models continue to exhibit significant performance disparities compared to their closed-source counterparts in medical applications. These open-source models enhance their performance through pre-training and fine-tuning on medical datasets\cite{OpenBioLLMs,pmc,biomistral,medalpaca}, with their performance ceiling dependent on the quality of the datasets, employed strategies, underlying models, and computational resources\cite{llmsurvey}. Despite pre-training and fine-tuning, the parameterized knowledge within the models remains insufficient to comprehensively address all medical issues, and the computational cost of retraining continues to be substantial\cite{nhzs}. When presented with novel medical queries, these models may generate responses that deviate from established medical knowledge, a phenomenon commonly referred to as "Hallucination"\cite{hallucination1,hallucination2}.

To address these challenges, Retrieval-Augmented Generation (RAG) technology enhances the knowledge base of models by retrieving information from external sources and incorporating it into the prompts of generative language models in a non-parametric manner\cite{RAGBZ,incontext}. This augmentation substantially improves performance in addressing domain-specific queries. Nevertheless, the integration of RAG techniques with LLMs presents several challenges, including the optimization of retrieval processes and their timing, effective utilization of contextual information, and mitigation of noise in retrieved documents\cite{ragsurvey1,ragsurvey2}. Research efforts have been initiated to address these challenges, encompassing adaptive retrieval methods (e.g., FLARE\cite{FLARE}), iterative retrieval (e.g., ITER-RETGEN\cite{RETGEN}), and fine-tuning adaptations for RAG tasks (e.g., Self-RAG\cite{selfrag} and RA-ISF\cite{isf}). Moreover, the integration of RAG with LLMs in the medical domain remains in its incipient stages\cite{almanac}. The strategic utilization of the strengths of both RAG and LLMs to bridge the performance gap between open-source and closed-source models remains an underexplored research avenue. Current research predominantly focuses on evaluating the impact of various components in RAG systems—including the LLMs, corpora, and retrieval models—through systematic assessment of their specific effects on performance\cite{benchmark}.

\begin{figure*}[ht!]
	\centering
	\includegraphics[width=400pt]{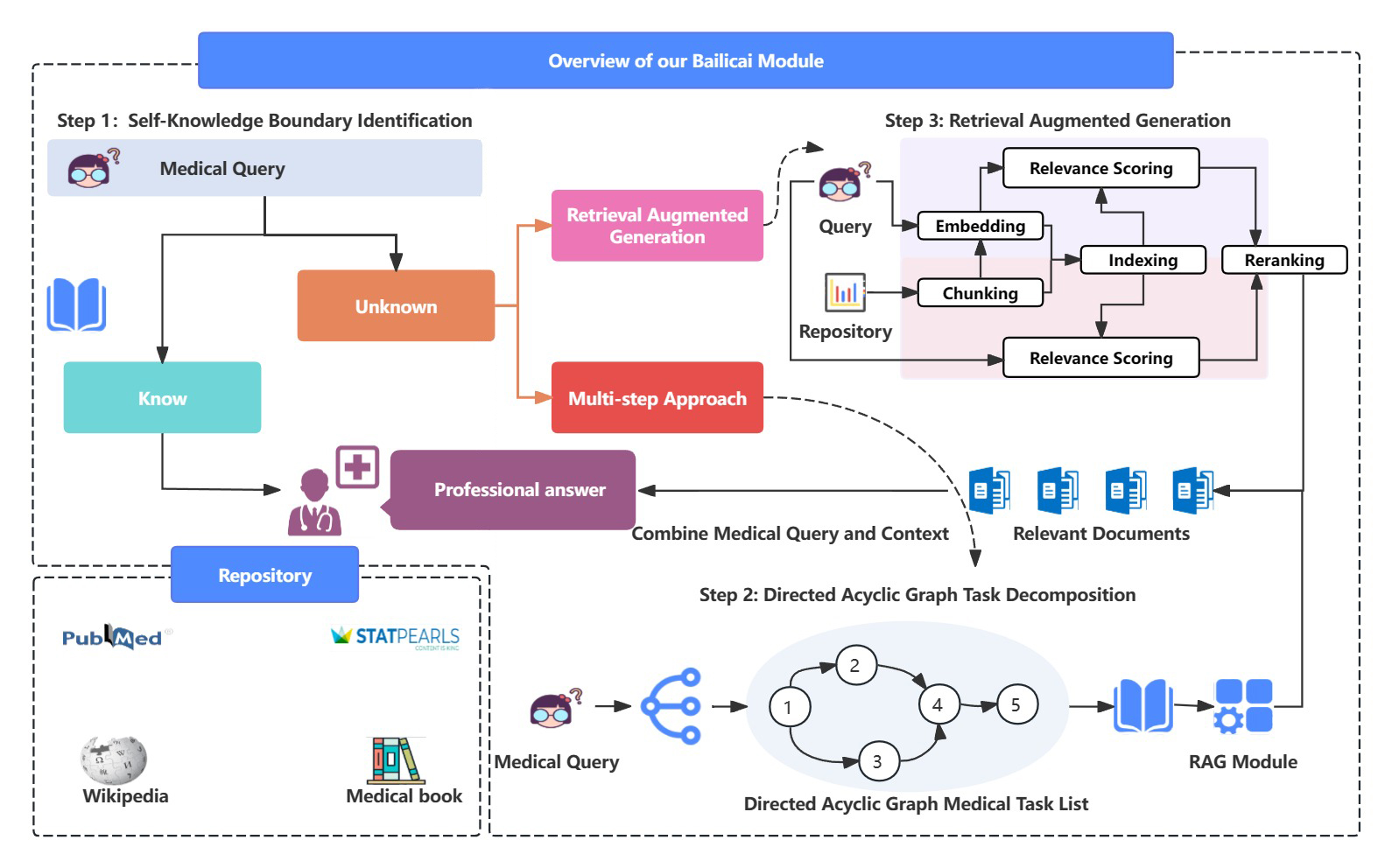} 
	\caption{Bailicai Overall Process Overview}
	\label{fig:myimage}
\end{figure*}

In response to these challenges, we propose the "Bailicai" framework, which efficiently integrates medical tasks with Retrieval-Augmented Generation (RAG) technology. By integrating RAG with open-source large language models, Bailicai achieves a synergistic effect that substantially improves performance in medical applications. The core architecture of the Bailicai framework, illustrated in Fig. 1, consists of four primary components: 1) Medical Knowledge Injection, 2) Self-Knowledge Boundary Identification, 3) Directed Acyclic Graph Task Decomposition, and 4) Retrieval-Augmented Generation. Initially, input queries are processed through the Self-Knowledge Boundary Identification module to determine the necessity of external knowledge. If external knowledge is not required, responses are generated directly via the Medical Knowledge Injection module. When external assistance is necessary, the query is directed to the Directed Acyclic Graph Task Decomposition module, which generates comprehensive contextual information for subsequent RAG execution. In the fine-tuning phase of Bailicai, we implemented a Model-Oriented Directive Data Filtering (MoDS\cite{mods}) method, based on previous research, to curate data from the UltraMedical dataset. This curated data forms the training foundation for the Bailicai model. Furthermore, we meticulously incorporated "hard negatives" into the literature-based question-answering sub-dataset, coupled with the MedPrompt prompting strategy to optimize sub-dataset construction. This approach ensures that the model's training focuses on key documents closely related to the query, while effectively filtering out irrelevant information. To enhance the integration of RAG technology, we abstracted the prerequisites for RAG into discrete components and developed corresponding datasets encompassing Self-Knowledge Boundary Identification and Directed Acyclic Graph Task Decomposition. Moreover, we employed prompting engineering techniques and MoDS for data distillation and selection on the Meta-Llama3-70B model\cite{llama3}, and applied LoRA\cite{lora} fine-tuning on the Meta-Llama3-8B model\cite{llama3}. In comparison with existing medical large language models and RAG approaches, Bailicai has exhibited superior performance across multiple medical benchmarks.

\textbf{Our principal contributions are as follows:}
\begin{itemize}
	\item To the best of our knowledge, this research presents the first implementation of a Directed Acyclic Graph Task Decomposition paradigm within the Retrieval-Augmented Generation framework, with specific application to the medical domain. This approach utilizes a hierarchical structure to systematically process subtasks, thus providing a more comprehensive and nuanced knowledge base for subsequent RAG applications. 
	\item We constructed the Bailicai dataset, integrating medical data from the UltraMedical dataset\cite{ultramedical} and incorporating high-quality data specifically curated for Self-Knowledge Boundary Identification and Directed Acyclic Graph Task Decomposition tasks. This dataset underwent a rigorous data filtering process to ensure optimal relevance and quality.
	\item The synergistic multi-module architecture of the Bailicai framework leverages the strengths of both large language models and Retrieval-Augmented Generation technology, substantially improving performance in medical applications. This integrated approach exhibits robust noise resistance and has been empirically validated through comprehensive experimentation.
\end{itemize}

\section{Related Work}
\label{s:Related Work}

\subsection{LLMs in Medical Domains}
\noindent
Since the introduction of GPT-3.5 in 2022\cite{gpt3.5}, large language models (LLMs) have exhibited breakthrough performance across diverse natural language processing (NLP) tasks\cite{openai,palm,llama3}. These models have been adapted for specific domains, producing promising results; notable examples include OceanGPT\cite{oceangpt} in marine science and LawGPT\cite{lawgpt} in legal studies.

In the medical domain, systems are increasingly integrating high-performance proprietary language models, such as OpenAI's GPT-4 and Google's PaLM, via API calls in conjunction with prompt engineering techniques. The efficacy of these applications has been validated in several instances, exemplified by GPT-3.5's performance on the United States Medical Licensing Examination (USMLE), where it has either met or approached the 60\% passing threshold\cite{exam}. MedPrompt, employing dynamic few-shot learning and self-generated reasoning chains, has enhanced GPT-4's accuracy on the USMLE to 90.2\%\cite{medprompt}. Utilizing comparable methodologies, the Med-PaLM model\cite{medpalm}, based on the PaLM architecture\cite{palm}, was the first to achieve a 67\% accuracy rate on the USMLE, with its subsequent iteration, Med-PaLM 2\cite{medpalm2}, attaining an accuracy of 86.5\%, commensurate with expert-level performance. However, the implementation of these proprietary models in medical applications raises concerns, particularly regarding potential privacy breaches\cite{llmmedicine}.

Concurrently, researchers are investigating the potential of open-source large language models, notably the Llama\cite{llama2,llama3} and Mistral\cite{mistral} series, by transforming medical corpora into question-answer pairs through pretraining and fine-tuning processes to enhance performance in the medical domain. Several models, including Med-Alpaca\cite{medalpaca}, BioMistral\cite{biomistral}, PMC-LLaMA\cite{pmc}, and OpenBioLLM\cite{OpenBioLLMs}, exemplify this approach. PMC-LLaMA, for example, incorporates an extensive corpus of medical papers and textbooks during its pretraining phase to encode domain-specific knowledge, followed by utilizing ChatGPT for instruction alignment during fine-tuning on a range of medical tasks. Notwithstanding these advancements, a significant performance disparity persists between open-source and proprietary models, attributable to the base model's capabilities, training methodologies, and dataset quality. Empirical evidence indicates that high-quality datasets yield more substantial performance improvements compared to larger volumes of lower-quality data\cite{lima}. Furthermore, computational resource constraints have necessitated the adoption of techniques such as low-rank approximation (LoRA)\cite{lora} to mitigate hardware demands during pretraining and fine-tuning phases.

\subsection{Retrieval-augmented Generation}
\noindent
Retrieval-Augmented Generation (RAG) enhances the performance of large language models by leveraging external knowledge bases to retrieve relevant document segments through semantic similarity computations\cite{RAGBZ,incontext}. This approach significantly reduces the incidence of hallucinations—defined as instances where generated content deviates from factual accuracy\cite{hallucination1,hallucination2}. Early research on RAG primarily focused on developing sparse or dense retrievers\cite{dpr,medcpt}, whereas contemporary studies have emphasized optimizing the integration of RAG with Large Language Models (LLMs). These optimizations encompass the timing of retrieval, methodological enhancements, and refined utilization of contextual information to mitigate noise within retrieved documents\cite{ragsurvey1,ragsurvey2}.

With respect to adaptive retrieval strategies, the FLARE project has proposed two novel methods: proactive retrieval based on retrieval instructions and confidence-based proactive retrieval, both designed to mitigate unnecessary retrievals\cite{FLARE}. Furthermore, process optimization has evolved from the "Rewrite-Retrieve-Read" paradigm\cite{rrr} to "ITER-RETGEN" implementing iterative retrieval to incrementally access more granular and comprehensive knowledge\cite{RETGEN}. Concerning the management of noise within retrieved documents, studies conducted demonstrated that irrelevant noise documents do not necessarily deteriorate system performance; conversely, they can enhance accuracy by up to 35\%. Conversely, documents incorrectly classified as relevant to the query introduce significant interference, substantially impacting the model's generative performance\cite{power}.

In the medical domain, the application of RAG technology remains in its nascent stages. The MEDRAG system assessed performance variations across diverse retrievers and corpora in medical question-answering tasks\cite{benchmark}. The Self-BioRAG\cite{biorag} project integrated Self-RAG\cite{selfrag} technology in medicine, optimizing it through reflective tokens that address retrieval timing, evaluate the relevance and supporting capacity of retrieved documents in answer generation, and assess the quality of generated outputs. Nevertheless, empirical evidence suggests that this approach fails to surpass the performance of models specifically optimized for medical datasets, potentially attributable to the limitations of smaller models in multi-task integration, as demonstrated by the AUTOACT study\cite{autoact}. Furthermore, during the fine-tuning phase, concurrent with data related to RAG tasks, it is imperative to augment the integration of medical knowledge to ensure the model's accurate recognition and utilization of this domain-specific information\cite{raft}.

\section{Method}
\label{s:Method}
\noindent
In the contemporary medical domain, the integration of retrieval-augmented generation (RAG) technology with large language models remains in its nascent stage. A significant proportion of research in this area remains exploratory. Identifying the optimal juncture for model-initiated retrieval through adaptive mechanisms presents challenges, while data noise in retrieved information can precipitate erroneous outputs during the generation process. Consequently, leveraging the extensive research of our predecessors, we propose Bailicai, a retrieval-augmented generation framework that has been rigorously refined and optimized at each stage of its development. This framework effectively mitigates data noise through the infusion of domain-specific medical knowledge, strategically modulates RAG invocation for knowledge augmentation via Self-knowledge Boundary Identification, and optimizes processes using Directed Acyclic Graph Task Decomposition, thus enhancing the efficacy of LLMs applications in the medical domain.

\subsection{Overview}
\noindent
As illustrated in Fig. 1, our model architecture comprises three fine-tuned large models and one module: $\mathcal{M}_{Medical}$, $\mathcal{M}_{Know}$, $\mathcal{M}_{DAG}$, and $\mathcal{M}_{RAG}$. Each component executes its designated function within the Bailicai framework, facilitating medical knowledge infusion, self-knowledge boundary identification, directed acyclic graph task decomposition, and retrieval-augmented generation. The framework's development process involves constructing task-specific datasets and optimizing the models for each respective task. During the training phase, synthetic noise was incorporated into the infusion of medical knowledge to augment the model's robustness in leveraging retrieval knowledge. In the ensuing reasoning phase, the process initiates with a user inputting a base query($\mathcal{Q}_{Base}$). $\mathcal{M}_{Know}$ initially evaluates whether the query can be addressed using the internal medical knowledge of $\mathcal{M}_{Medical}$, thereby determining the necessity for external retrieval. If addressable, the answer $\mathcal{A}_{Fin}$ is generated directly. If not, the query is processed sequentially through $\mathcal{M}_{DAG}$ and $\mathcal{M}_{RAG}$ to decompose the question, retrieve, and organize the relevant information. Subsequently, the retrieved information $\mathcal{K}_{RAG}$, in conjunction with the answer paragraph $\mathcal{K}_{Dirt}$ (enriched by $\mathcal{M}_{Medical}$) and the sub-question and its answer $\mathcal{K}_{DAG}$, are integrated to form comprehensive background information. This integrated composite serves as a new prompt input for $\mathcal{M}_{Medical}$ to generate the final answer $\mathcal{A}_{Fin}$.

\subsection{Self-Knowledge Boundary Identification}
\noindent
Existing retrieval-augmented generation (RAG) techniques predominantly enhance the input of large language models by incorporating relevant textual segments, aiming to mitigate factual errors in knowledge-intensive tasks. However, this approach exhibits excessive reliance on the efficacy of the retrieval module and fails to fully leverage the inherent parameterized knowledge of the large language model, potentially compromising output quality. Furthermore, while these techniques may augment task performance, they concurrently introduce significant processing latency, potentially limiting their applicability in real-time scenarios.

Within the Bailicai framework, the $\mathcal{M}_{Know}$ model conducts an initial assessment to determine if the user's presented medical query can be resolved utilizing the internal parameterized knowledge base of the foundational model, $\mathcal{M}_{Medical}$. This process delineates the aspects of the query that are sufficiently addressed by the extant knowledge and those that necessitate supplementary information retrieval or problem decomposition. This methodological approach precisely identifies the scope, timing, and depth of requisite information retrieval, thereby substantially mitigating the temporal overhead associated with comprehensive retrieval processes.

If the inference outcome of the $\mathcal{M}_{Know}$ model is classified as "know", the query is promptly directed to the $\mathcal{M}_{Medical}$ model, which subsequently generates the initial response, $\mathcal{A}_{Base}$. Conversely, if the inference outcome is categorized as "unknow", the query advances to the second stage: Directed Acyclic Graph Task Decomposition. In this stage, the query undergoes further decomposition to facilitate subsequent information retrieval and processing. The formalized representation is delineated as follows:
\begin{equation}
	\begin{aligned}
		\mathcal{M}_{Know}({Q}_{Base}) &= \arg \max_{a}P(a|\mathcal{P}_{Know}(\mathcal{Q}_{Base}), \\
		&\qquad \qquad \qquad \qquad \quad \;\mathcal{M}_{Know})
	\end{aligned}
\end{equation}

\subsection{Directed Acyclic Graph Task Decomposition}
\noindent
In the medical domain, addressing complex clinical diagnostic tasks requires meticulous planning of diagnostic protocols, systematic assessment of symptoms against established diagnostic criteria, and implementation of appropriate management strategies. Extant research on task decomposition predominantly employs prompt engineering methodologies (e.g., Chains of Thought\cite{cot}) to partition tasks into multiple fine-grained subtasks. Through the resolution of these subtasks, requisite information is provided to the primary task, thereby augmenting the quality of the final outcome. However, this task decomposition model may face limitations in effectively articulating and managing complex dependencies inherent in medical tasks. Inspired by research on data science planning\cite{dag}, we implement a hierarchical structure of directed acyclic graphs to enhance comprehension and management of the complexity inherent in medical tasks. This structure enables a precise representation of inter-task dependencies, facilitating more systematic and efficient task processing. Furthermore, the graph structure incorporates metadata for each task node, comprising a detailed task description, completion status, antecedent and subsequent dependencies, and achieved outcomes. The hierarchical structure of directed acyclic graphs elucidates the mutual influences among tasks while concurrently enhancing the efficiency and precision in the management of complex medical tasks.

We leverage the language generation capabilities of large language models to decompose complex tasks and represent them via directed acyclic graphs (DAGs). These graph structures efficaciously elucidate the dependencies inherent in task execution. Specifically, we convey the fundamental question $\mathcal{Q}_{Base}$ to $\mathcal{M}_{DAG}$ by initializing $\mathcal{P}_{DAG}$ combinations, subsequently generating a directed acyclic graph question list $\mathcal Q_{DAG} = \{\mathcal Q_{1}, \mathcal Q_{2}, ..., \mathcal Q_{n}\}$. Detailed formulas are provided below to elucidate the methodology:
\begin{equation}
	\mathcal{Q}_{DAG} = \arg \max_{a}P(a|\mathcal{P}_{DAG}(\mathcal{Q}_{Base}),\; \mathcal{M}_{DAG})
\end{equation}

Subsequently, we reintegrate the sub-questions into the model $\mathcal{M}_{Know}$ to dynamically evaluate the necessity of invoking the external retrieval knowledge module $\mathcal{M}_{RAG}$ for knowledge augmentation. 

When $\mathcal{M}_{Know}(\mathcal{Q}_i)$ = know, the model can address the current question utilizing its intrinsic medical knowledge. We define $\mathcal{A}_{i}$ as follows:
\begin{equation}
	\mathcal{A}_i = \mathcal{M}_{Medical}(\mathcal{Q}_i)
\end{equation}

When $\mathcal{M}_{Know}(\mathcal{Q}_i)$ = unknown, meaning the model needs to obtain external knowledge to answer, we define $\mathcal{A}_{i}$ as follows:
\begin{equation}
	\mathcal{A}_i = \mathcal{M}_{Medical}(\mathcal M_{RAG}(\mathcal{Q}_i) \oplus \widehat{A} \oplus \mathcal{Q}_i)
\end{equation}

Among them, the answer $\widehat{A}$ corresponds to the set of previous tasks upon which question $\mathcal{Q}_i$ depends.
 
In this study, we employed the Meta-Llama-3-70B model to conduct task-driven adaptation on a specialized medical instruction dataset. Initially, we acquired and curated relevant data in accordance with predefined prompts (see Table 1) for preliminary knowledge distillation. Subsequently, we implemented a rigorous data cleaning process to eliminate inconsistent and redundant data.

\subsection{Medical Knowledge Injection}
\noindent
Open-source large language models demonstrate the potential to attain expert-level performance in specialized medical tasks through supervised fine-tuning on domain-specific datasets. Conversely, research on integrating Retrieval-Augmented Generation (RAG) with large language models for medical applications remains in its nascent stages. Our findings indicate that the performance enhancement resulting from this integration is contingent upon the intrinsic capabilities of the large model and the domain-specific requirements. Inspired by previous research\cite{raft}, we deliberately incorporated golden documents ($\mathcal{D}_*$) and interference documents ($\mathcal{D}_-$) during training to augment the model's capacity to accurately cite golden documents and disregard interference documents in tasks involving medical literature. For interference documents ($\mathcal{D}_-$), we employ hard negatives to identify documents within the dataset. We classified medical tasks into three categories: medical examination questions, literature-based question answering, and open-ended queries, to comprehensively evaluate the efficacy of RAG in these domains. The methodological details are delineated as follows:
\begin{equation}
	\mathcal D_{Medical} : \mathcal{Q}_{Base} + \mathcal{D}_{*} + \mathcal{D}_{-} \rightarrow \mathcal{A}_{Fin} 
\end{equation}

During the ensuing inference phase, the conceptual framework can be expressed as follows:
\begin{equation}
	\mathcal{A}_{Fin} = \arg \max_{a}P(a|\mathcal{Q},\; \mathcal{M}_{Medical})
\end{equation}

\subsection{Retrieval-Augmented Generation}
\noindent
In the domains of biomedicine and clinical practice, researchers and clinicians frequently leverage supplementary information to address complex problems. Analogously, language models augment their problem-solving capabilities through the retrieval of pertinent documents. This study utilizes the MedCPT model for retrieval. During the construction phase, we implement a dense encoder, $E_{P}$(·), which maps text paragraphs to d-dimensional real vectors and generates an index database comprising $M$ paragraphs for retrieval. In the query process, another encoder, $E_{Q}$(·), transforms the input question into a d-dimensional vector, which is subsequently utilized to retrieve the $K$ paragraphs exhibiting maximum similarity to the question vector. The similarity between the question vector and document vectors is computed via the dot product. The detailed implementation methodology is delineated as follows:
\begin{equation}
	\operatorname{sim}(\mathcal{Q}_{Base}, \mathcal{P}_{RAG}) = E_Q(\mathcal{Q}_{Base})^\intercal E_P(\mathcal{P}_{RAG})	
\end{equation}

\section{Experimental}
\label{s:Experimental}
\subsection{The Bailicai Medical Dataset}
\noindent
The Bailicai Medical Dataset comprises both a training dataset and a retrieval dataset. The training dataset, extracted from the UltraMedical Dataset, encompasses approximately 277,677 high-quality entries of medical instruction data. UltraMedical incorporates over 410,000 meticulously curated medical instructions, integrating both manually crafted and algorithmically-generated synthetic prompts. The construction of UltraMedical involves the extraction of high-quality data from medical examination databases and PubMed articles, supplemented by synthetic prompts generated through sophisticated algorithms, including self-evolution and heuristic-based filtering\cite{ultramedical}. The medical instruction data illustrated in Fig. 2 will be elaborated upon:

\begin{figure*}[ht!]
	\centering
	\includegraphics[width=400pt]{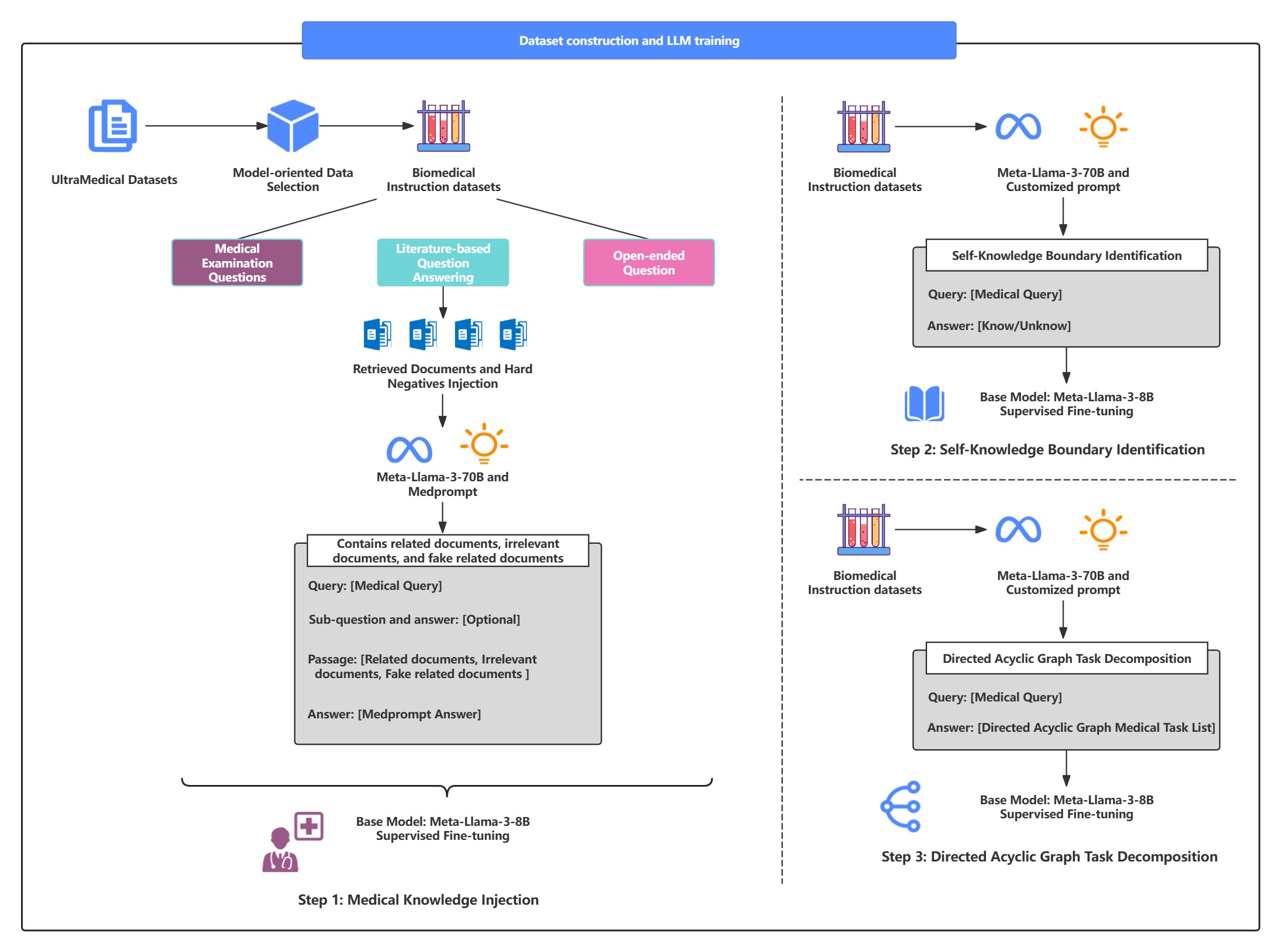} 
	\caption{Bailicai dataset construction and LLM training}
	\label{fig:myimage}
\end{figure*}

Recent studies have elucidated that during the instruction execution phase, a limited quantity of high-quality data typically yields superior performance compared to a large volume of low-quality data. Considering the constraints of computational resources, we commenced our analysis with the UltraMedical dataset, utilizing the model's score on the original data to establish a threshold of 9 points for initial screening. Subsequently, we employed the Model-Oriented Instruction Data Selection method (MODS\cite{mods}) to further optimize the dataset. Specifically, we utilized OpenAssistant's reward-model-debertav3-large-v2 to evaluate the data quality. This model, based on the DeBERTa architecture, facilitated the further refinement of high-quality instruction data through the implementation of a score threshold of 1. Additionally, we implemented the k-center greedy algorithm to select a subset of data points exhibiting maximal inter-point distances, thereby ensuring diversity and comprehensive coverage in the retained dataset. Ultimately, 173,271 records were preserved as the core dataset for Bailicai.

We delineate the medical domain into three distinct tasks: self-knowledge boundary identification, directed acyclic graph task decomposition, and medical knowledge injection. These tasks are further refined and operationalized for specific applications through manual and large language model-based prompt engineering. Subsequently, the original dataset is filtered based on three criteria: data quality, coverage, and necessity. 

\textbf{1)\;Self-Knowledge Boundary Identification} A random subset of questions was extracted from the baseline dataset, and the Meta-Llama-3-70B model was utilized to perform task-driven adaptation on this subset. Relevant data were aggregated based on predefined prompts (see Table 1), followed by preliminary knowledge distillation. Subsequently, a data cleaning process was implemented to eliminate inconsistencies and redundant information from the dataset, incorporating manual sampling scores obtained using GPT-4. Upon completion of this methodology, the resulting dataset comprised 71,557 data points.
 
\begin{figure*}[ht!] 
	\centering 
	\begin{tcolorbox}[
		enhanced,
		width=\textwidth, 
		center, 
		colback=white, 
		colframe=black, 
		boxsep=5pt, 
		left=5pt, 
		right=5pt, 
		top=5pt, 
		bottom=5pt, 
		sharp corners=north, 
		rounded corners=south, 
		]
		=========================================\textbf{Prompt}=========================================
		Below is an instruction that describes a task. Write a response that appropriately completes the request.
		\\
		\\
		\#\#\# Instruction:
		\\
		Based on your inner knowledge, determine whether you can answer this medical question. If you are sure you can answer it, please use "know" as your only answer. If you cannot answer it or are not confident about it, please use "unknow" as your only answer. If the actual situation is that you cannot answer it or the answer is incorrect, but you say "know", you will be severely punished.\\\\
		Medical question: \{Query\}\\\\
		\#\#\# Response:
	\end{tcolorbox}
	\captionof{table}{Prompt template for Self-Knowledge Boundary Identification with Bailicai} 
\end{figure*}

\textbf{2)\;Directed Acyclic Graph Task Decomposition} Similarly, we adhered to the preset improvement protocol (see Table 3). Following this series of steps, the resulting dataset comprised 32,849 data points.

\textbf{3)\;Medical Knowledge Injection} The model's training efficacy was enhanced by incorporating distractor documents into the baseline literature-based question-answering dataset and implementing the MedPrompt prompt strategy. The thought chain generation in GPT-4 was refined using techniques such as similarity-based Dynamic Few-shot Learning, Self-Generated Chain of Thought, and Choice Shuffling Ensemble. These techniques facilitate more comprehensive thought chain generation and improve the model's positional accuracy in processing multiple-choice question answers, thus enhancing response quality. Moreover, this approach enables the model to focus more effectively on question-relevant documents while disregarding extraneous information. The resulting dataset after these refinements comprised 173,271 data points.

\textbf{4)\;Retrieval-Augmented Generation} Raw data were collected from four distinct sources to enhance retrieval generation.  The sources included PubMed (biomedical abstracts), Wikipedia (online encyclopedia), StatPearls (medical education database), and medical textbooks.  Furthermore, MedCPT was employed to train an integrated retrieval and re-ranking model using 255 million query records from PubMed search logs, utilizing a comparative training method.  Text paragraphs were transformed into vectors and stored in the Faiss vector database, facilitating efficient subsequent retrieval.  Table 2 provides comprehensive statistics regarding the medical retrieval corpus and the number of indexed documents.

\begin{table}[htb]
	\setlength{\belowcaptionskip}{0cm}
	\centering
	\caption{Characteristics of the Indexed Medical Corpus. Note: The 'Embedding' denotes the file size resulting from processing with the MedCPT model.}	
	\rowcolors{1}{gray!18}{white}
	\begin{tabular}{p{1.15cm}>{\centering\arraybackslash}p{1.15cm}>{\centering\arraybackslash}p{1.35cm}l} 
		\toprule
		Data & Chunks & Embedding & Introduction \\ 
		\midrule
		PubMed & 23.9M & 80.1G & Medical Literature \\
		Wikipedia & 29.9M & 100.3G & General Knowledge \\
		StatPearls & 301.2K & 900M & Medical Database \\
		Textbooks & 125.8K & 368.7M & Medical Education \\
		Merge & 54.2M & 181.7G & Knowledge Combination \\
		\bottomrule
	\end{tabular}
	\label{tab:example_table}
\end{table}

\subsection{Training Details}
\noindent 
This study employs the Bailicai dataset, which is processed to conform to the Alpaca dataset format for fine-tuning on various tasks. Supervised fine-tuning (SFT) was conducted on the Meta-Llama-3-8B model using four NVIDIA Tesla V100 graphics processing units (GPUs), implementing the low-rank adaptation (LoRA) method. LoRA utilizes a low-rank decomposition matrix to approximate parameter updates for each network layer, potentially minimizing the number of parameters required for training when adapting to downstream tasks \cite{lora}. The cross-entropy loss function was employed to calculate the loss value during the fine-tuning process. The maximum context lengths for models $\mathcal{M}_{Know}$, $\mathcal{M}_{DAG}$, and $\mathcal{M}_{Medical}$ were configured to 1024, 2048, and 2816 tokens, respectively. All three models employed the AdamW optimizer and underwent domain-specific fine-tuning in the medical field for three epochs, with a batch size of 2 and a learning rate of 2e-5. Furthermore, the LoRA hyperparameters r, alpha, and dropout were set to 128, 256, and 0.1, respectively.

\begin{figure*}[ht!] 
	\centering 
	\begin{tcolorbox}[
		enhanced,
		width=\textwidth, 
		center, 
		colback=white, 
		colframe=black, 
		boxsep=5pt, 
		left=5pt, 
		right=5pt, 
		top=5pt, 
		bottom=5pt, 
		sharp corners=north, 
		rounded corners=south, 
		]
		=========================================\textbf{Prompt}=========================================
		Below is an instruction that describes a task. Write a response that appropriately completes the request.
		\\
		\\
		\#\#\# Instruction:
		
		Please analyze the medical question provided and try to break the question into smaller, distinct sub-questions. By subsequently solving the sub-problems and combining the answers to the sub-problems, the original question can be better answered. A problem consists of one to four sub-problems.
		\\
		\\
		When posing questions based on patient symptom data, all sub-questions should be expanded from the given options to include corresponding background information, rather than directly asking questions based on the patient's symptom data.
		If the problem can be decomposed, carefully follow the instruction, don't make unnecessary changes and do not answer any sub-questions.
		\\
		\\
		Output a list of Json following the format:
\begin{verbatim}
[
 {
"task_id": "unique identifier for a sub-problem, can be an ordinal",
"dependent_task_ids": ["The sub-issue ID of the prerequisite for this sub-issue"],
"instruction": "what you should do in this sub-problem, one short phrase or sentence
 "},...
]
\end{verbatim}
		If the problem does not need to be broken down into sub-problems, your output format is:
\begin{verbatim}
[
 {
    "task_id": "1",
    "dependent_task_ids": [],
    "instruction": original question
 }
]
\end{verbatim}	
		Medical question: \{Query\}\\\\
		\#\#\# Response:
	\end{tcolorbox}
	\captionof{table}{Prompt template for Directed Acyclic Graph Task Decomposition with Bailicai} 
\end{figure*}

\subsection{Retrieval-Augmented Generation}
\noindent
The Retrieval-Augmented Generation (RAG) module employs a two-stage retrieval strategy, consisting of coarse-grained and fine-grained retrieval phases.

The coarse-grained retrieval stage initiates with information extraction from four medically-related corpora. MedCPT, a unified retrieval tool, vectorizes all texts in the corpus. To optimize retrieval efficiency for large-scale data, a Hierarchical Navigable Small World (HNSW) index is implemented within the Faiss vector database. The retrieval process commences at the top layer, utilizing a greedy algorithm to identify the node closest to the query vector. Subsequently, the search progresses downward, exploring neighboring nodes at each layer until reaching the bottom layer.

The fine-grained retrieval stage employs a reranking model (bge-reranker-large) to score the top 32 retrieved paragraphs. The scoring criteria encompass semantic consistency, answer completeness, logical coherence, and relevance to the user query. Ultimately, the five highest-scoring documents are selected for further processing.

\subsection{Medical Domain Benchmarks}
\noindent
To comprehensively evaluate the domain expertise of the Bailicai-based large language model in the medical field and its retrieval augmentation efficacy, this study utilized five medical question-answering benchmark datasets\cite{medpalm,benchmark}: MedQA, MedMCQA, MMLU-Med, PubMedQA, and BioASQ. MMLU-Med, a subset of the MMLU's medical category, comprises tasks in anatomy, clinical knowledge, university medicine, medical genetics, professional medicine, and university biology. The study evaluates the model's overall performance in medical question-answering for each dataset and task by computing its predictive accuracy. Detailed characteristics of the specific datasets are presented in Table 4.

\begin{table*}[htbp]
	\setlength{\belowcaptionskip}{0cm}
	\centering
	\caption{Descriptive Statistics of Medical Domain Benchmarks}
	\begin{tabular*}{\textwidth}{@{\extracolsep{\fill}} lccc p{5cm}c}
		\toprule
		\bfseries Dataset & \bfseries Format & \bfseries Options & \bfseries Size & \bfseries A concise description & \bfseries Source\\
		\midrule
		MedQA & Question + Answer & A / B /C / D & 1273 & United States medical licensing & Examination \\
		& & & & examination & \\
		\midrule
		MedMCQA & Question + Answer & A / B / C / D & 4183 & Medical entrance examinations in&  Examination \\
		& + Explanations & & & India & \\
		\midrule
		MMLU & Question + Answer & A / B / C / D & 1089 & Encompasses biomedical knowledge, including anatomy, clinical knowledge, and professional medicine, etc &  Examination \\
		\midrule
		BioASQ & Question + Answer & Yes / No & 618 & Biomedical semantic indexing and question answering systems encompass a wide range of medical subfields & Literature\\
		\midrule
		PubMedQA & Question + Context & Yes / No / Maybe & 500 &  Biomedical Literature Question & Literature\\
		& + Answer & & & Answering & \\
		\bottomrule
	\end{tabular*}
	\label{tab:qa_datasets}
\end{table*}

\subsection{Baselines}
\noindent
To comprehensively assess and compare the Bailicai framework, benchmarks were conducted utilizing a standardized medical evaluation protocol, retrieval model, and a domain-specific medical retrieval corpus. The study encompassed baseline models and methods from three domains: general-purpose, domain-specific, and Retrieval-Augmented Generation (RAG). 

\textbf{1)\;Chain-of-Thought (CoT)\cite{cot}} This method facilitates step-by-step reasoning through prompts such as "Let's think step by step," thereby enhancing problem-solving accuracy.

\textbf{2)\;Med-Alpaca\cite{medalpaca}} This model is fine-tuned on Llama using a curated dataset of over 160,000 medical domain-specific entries, comprising medical literature and authentic Q\&A pairs.

\textbf{3)\;BioMistral\cite{biomistral}} This model, based on the Mistral architecture, undergoes additional pre-training on an extensive corpus of medical literature from PubMed Central to enhance its proficiency in medical terminology and question-handling. It is subsequently fine-tuned on the MedQA, PubMedQA, and MedMCQA datasets.

\textbf{4)\;PMC-LLaMA\cite{pmc}} This model incorporates a comprehensive corpus comprising 4.8 million biomedical academic papers and 30,000 medical textbooks, enhancing its medical knowledge integration. Furthermore, it undergoes domain-specific instructional tuning on a 202-million-token dataset, encompassing medical Q\&A and knowledge graphs.

\textbf{5)\;OpenBioLLM\cite{OpenBioLLMs}}  Fine-tuned on Meta-Llama-3-8B using a comprehensive medical instruction dataset, this model incorporates Direct Preference Optimization during fine-tuning, achieving high performance in multi-task scenarios including clinical note summarization and medical Q\&A.

\textbf{6)\;Flan-PaLM\cite{medpalm}} This variant of the 540-billion parameter PaLM model is fine-tuned on MultiMedQA, utilizing strategies such as few-shot prompts, Chain-of-Thought, and consistency prompts. It demonstrates state-of-the-art performance across multiple medical Q\&A datasets.

\textbf{7)\;Med-PaLM2\cite{medpalm2}} This model integrates an enhanced foundational LLMs (PaLM 2), domain-specific fine-tuning, and a novel Ensemble Refinement prompting strategy. It effectively bridges the gap with clinical responses, demonstrating significant improvements across various medical benchmarks.

\textbf{8)\;Mistral-7B-V0.3\cite{mistral}} MistralAI's third-generation model expands the vocabulary and introduces new functionalities, such as function calls, compared to its predecessors.

\textbf{9)\;Meta-Llama-3\cite{llama3}} Meta's third-generation model significantly expands its vocabulary and implements Grouped Query Attention (GQA). It employs diverse filtering technologies, including heuristic filters, NSFW filters, and semantic deduplication, to enhance training data quality. The model also incorporates advanced techniques such as Supervised Fine-Tuning (SFT), rejection sampling, Proximal Policy Optimization (PPO), and Direct Preference Optimization (DPO).

\textbf{10)\;ChatGPT\cite{openai}} Launched by OpenAI in November 2022, this commercial model demonstrates superior performance across diverse domains and natural language processing tasks.

\textbf{11)\;Self-BioRAG\cite{biorag}} This model adapts the Self-RAG framework to the medical domain, utilizing GPT-4 to generate specific reflection markers for knowledge distillation. It manages various aspects of the retrieval process, including retrieval decisions and content relevance assessment.

\textbf{12)\;MEDRAG\cite{benchmark}} This approach integrates multiple domain-specific medical corpora and conducts comprehensive evaluations across various retrieval models and large language models.

\section{Results and Discussion}
\noindent
This study presents a series of experiments designed to assess the performance of the Bailicai model on medical domain tasks in comparison with baseline models. The evaluation encompassed comparisons between Bailicai and models specifically fine-tuned for the medical domain, as well as models and frameworks optimized for retrieval-augmented generation tasks. Experimental findings indicate that Bailicai exhibits substantial advantages in processing medical-related tasks. Furthermore, a series of ablation studies was conducted to quantify the specific impact of various components on the model's overall performance.

\begin{table*}[htb]
	\setlength{\belowcaptionskip}{0cm}
	\centering
	\caption{\textbf{Primary Experimental Outcomes.} Each baseline model was evaluated under standardized settings. The experimental protocol consistently utilized a chain-of-thought strategy in prompts, with the Temperature parameter maintained at 0. To retrieval-augmented generation, both the MedCPT model and the PubMed knowledge base are utilized. \textbf{Bold} values denote the superior performance among all methods for each model. \textbf{*} signify results reported in the original paper.}
	\label{tab:methods}
	\begin{tabular}{@{}lccccccc@{}}
		\toprule
		\bfseries Model & \bfseries Parameter & \bfseries MedQA & \bfseries MedMCQA & \bfseries MMLU-Med & \bfseries PubMedQA & \bfseries BioASQ & \bfseries Average \\
		\midrule
		\multicolumn{8}{c}{\bfseries Medical Large Language Model (COT)} \\
		\midrule 
		MedAlpaca & 7B & 37.63 & 32.7 & 57.58 & 49.8 & 62.62 &  48.07 \\
		BioMistral & 7B & 43.21 & 41.33 & 59.32 & 42.00 & 70.23 & 51.24 \\
		PMC-LLaMA & 13B & 56.40 & 47.55 & 51.88 & 61.60 & 66.34 & 56.75 \\
		OpenBioLLM & 8B & 57.11 & 56.03 & 75.11 & 71.80 & 59.87 & 63.98 \\
		Flan-PaLM* & 540B & 67.60 & 57.60 & 79.16 & 79.00 & - & 70.84 \\
		Med-PaLM2* & 340B & 83.70 & 71.50 & 88.43 & 74.00 & - & 79.41 \\
		\midrule
		\multicolumn{8}{c}{\bfseries Universal Large Language Model (COT)} \\
		\midrule 
		Mistral-7B-v0.3 & 7B & 48.78 & 45.35 & 63.64 & 48.00 & 68.93 & 54.94 \\
		Meta-Llama-3 & 8B & 62.69 & 52.52 & 72.73 & 75.40 & 72.01 & 67.07 \\
		Meta-Llama-3 & 70B & 76.98 & 65.77 & 84.30 & 76.80 & 79.13 & 76.60 \\
		ChatGPT-3.5* & - & 57.71 & 53.79 & 70.84 & 72.66 & 74.27 & 65.85 \\
		ChatGPT-4* & - & 78.80 & 69.50 & 87.20 & 70.00 & 84.30 & 79.00 \\
		\midrule
		\multicolumn{8}{c}{\bfseries Retrieval-Augmented Generation} \\
		\midrule 
		Meta-Llama-3-RAG & 8B & 57.97 & 55.27 & 70.71 & 76.40 & 81.55 & 68.38 \\
		Self-BioRAG\* & 13B & 48.60 & 44.00 & 57.20 & 54.60 & - & 51.10 \\
		ChatGPT-3.5-MEDRAG* & - & 66.61 & 58.04 & 75.48 & 67.40 & 90.29 & 71.56 \\
		ChatGPT-4-MEDRAG* & - & 82.80 & 66.65 & 87.24 & 70.60 & 92.56 & 79.97 \\
		\midrule
		\textbf{Our Model} \\
		\midrule 
		\bfseries Bailicai & \bfseries 8B &\bfseries 66.85 &  \bfseries55.89 & \bfseries 76.12 &  \bfseries77.40 & \bfseries 82.85 & \bfseries 71.82 \\
		\bottomrule
	\end{tabular}
\end{table*}

\subsection{Main Results}
\textbf{Comparison with Large Models in the Medical Domain} Table 5 demonstrates the superior performance of the Bailicai model across five medical benchmarks. In comparison with domain-specific models of smaller scale (e.g., 7B, 8B, 13B parameters) that are pre-trained and fine-tuned for medical tasks, Bailicai attains state-of-the-art results on benchmarks including MedQA, MMLU-Med, PubMedQA, and BioASQ. On the MedMCQA benchmark, Bailicai's performance is comparable to OpenBioLLM, with a marginal difference of 0.14\% and an average score of 71.82\%. Moreover, Bailicai (8B) demonstrates performance parity with larger models such as Flan-Palm (540B) in the medical domain, a characteristic of particular importance in resource-constrained healthcare environments due to its capability for local deployment coupled with the absence of privacy leakage concerns. These findings underscore Bailicai's substantial advantages in processing diverse medical tasks.

\textbf{Comparison with General Domain Large Models} As evidenced in Table 5, Bailicai outperforms ChatGPT-3.5(achieving an average score of 65.85\%) by 5.97\%. In the medical domain, given the critical nature of user privacy, Bailicai offers a significant advantage through its capability for local deployment. Furthermore, Bailicai achieves superior performance relative to the Meta-Llama-3-70B model on the PubMedQA and BioASQ benchmarks, surpassing it by 0.6\% and 3.72\% respectively.

\textbf{Comparison with Retrieval-Augmented Generation Models and Methods} Tables 5 and 6 illustrate that Bailicai effectively mitigates hallucination issues stemming from noise in retrieved documents through retrieval-enhanced generation, outperforming other retrieval-augmented generation models.  Notably, in comparison to the Self-BioRAG model, which is specifically fine-tuned for retrieval-augmented generation, Bailicai demonstrates a significant advantage across multiple datasets, yielding an average performance improvement of 20.72\%.

\begin{table*}[htbp]
	\setlength{\belowcaptionskip}{0cm}
	\centering
	\caption{\textbf{Ablation Study Outcomes: Analysis of Bailicai Component Efficacy.} To enhance retrieval performance, both the MedCPT model and the PubMed knowledge base are utilized. \textbf{Bold} values denote the superior performance among all methods for each model.}
	\begin{tabular}{lcccccc}
		\toprule
		\bfseries Experiment Detail & \bfseries MedQA &\bfseries  MedMCQA & \bfseries MMUL-Med & \bfseries PubMedQA & \bfseries BioASQ &\bfseries  Average \\
		\midrule
		Meta-Llama-3-8B & 62.69 & 52.52 & 72.73 & 75.40 & 72.01 & 67.07 \\
		\; + Retrieval-Augmented Generation & 57.97 & 55.27 & 70.71 & 76.40 & 81.55 & 68.38 \\
		Bailicai \\
		\; + Medical Knowledge Injection &65.67 & 55.82 & 75.85 & 77.40 & 78.80 & 70.71\\
		\quad + Self-Knowledge Boundary\\
		\quad\; + Retrieval-Augmented Generation &66.06 & \bfseries 56.08 & 75.67& \bfseries 78.40 & 81.72& 71.59\\
		\quad\; + DAG Task Decomposition &65.75 & 55.85 & 75.85 & 77.20 & 80.23 & 70.98 \\
		\quad\quad \bfseries+ Retrieval-Augmented Generation &\bfseries 66.85 &  55.89 & \bfseries 76.12 &  77.40 & \bfseries 82.85 & \bfseries 71.82 \\
		\bottomrule
	\end{tabular}
	\label{tab:medical_qa}
\end{table*}

\subsection{Ablation Studies}
\noindent
To evaluate the efficacy of various components within the Bailicai framework, a series of ablation studies was conducted, as detailed in Table 6. Each experiment incorporated the integration of distinct modules:

\textbf{1)\;Medical Knowledge Injection}: This variant implements domain-specific fine-tuning on the Bailicai medical dataset, encompassing medical examinations, literature-based research, and open-domain medical question answering. The dataset comprises 17,000 entries, utilizing Meta-Llama3-8b as the underlying model.

\textbf{2)\;Self-Knowledge Boundary Identification}: Building upon domain fine-tuning, this variant integrates a self-knowledge boundary identification module. This module determines whether a query can be resolved solely using the medical knowledge parameterized within the model.

\textbf{3)\;Directed Acyclic Graph Task Decomposition}: This variant enhances the model's capacity to process complex queries by decomposing them into a sequence of structured sub-tasks. This approach addresses questions that cannot be resolved exclusively through internal or external knowledge.

\textbf{4)\;Retrieval-Augmented Generation}: This variant incorporates RAG technology to manage unknown or challenging queries. The process entails retrieving relevant documents and embedding them alongside the query in the prompt, thereby augmenting external knowledge.

Table 6 delineates the results of ablation studies, revealing that the removal of any module from the Bailicai framework leads to a performance decline, thus highlighting the critical contributions of each module to the overall efficacy. In comparison to the baseline model, Meta-Llama-3-8B, which utilizes Retrieval-Augmented Generation technology, the Bailicai framework demonstrates superior performance in information extraction and increased robustness against distracting documents. Concurrent application of all four modules yields performance improvements of 8.88\% and 5.41\% on the MedQA and MMLU-Med benchmarks, respectively, with an average score increase of 3.44\%.

Experimental results indicate that the integration of Retrieval-Augmented Generation technology with the Directed Acyclic Graph Task Decomposition module optimizes overall performance, achieving 71.82\%. This combined approach results in a marginal 0.23\% increase in overall performance compared to the isolated use of Retrieval-Augmented Generation technology, while concurrently producing a 1.0\% performance decline on the PubMedQA benchmark. This performance decrease may be attributed to the inherent structure of the PubMedQA dataset, which incorporates key contextual documents essential for question answering. Document retrieval following task decomposition potentially generates token lengths exceeding the maximum threshold (2812) established during the Medical Knowledge Injection phase of Bailicai, consequently impacting performance negatively.

To test this hypothesis, the experiments on the PubMedQA benchmark were reconfigured, excluding the golden context documents provided in the original dataset and focusing exclusively on question-based evaluation.  As evidenced in Table 7, the incremental addition of modules yielded an improvement in Bailicai's performance on the PubMedQA benchmark from 52.60\% to 69.80\%, corroborating the hypothesized causes of the initial performance decline.

\begin{table}[htb]
	\setlength{\belowcaptionskip}{0cm}
	\centering
	\caption{Comparative Analysis of Bailicai Component Performance on the PubMedQA*.   PubMedQA*: excludes original golden documents, retaining only the query and answer.}	
	\rowcolors{1}{gray!18}{white}
	\begin{tabular}{lc} 
		\toprule
		Module & PubMedQA*\\ 
		\midrule
		+ Medical Knowledge Injection & 52.60\\
		\;+ DAG Task Decomposition & 53.80\\
		\quad + Retrieval-Augmented Generation & 69.80\\
		\bottomrule
	\end{tabular}
	\label{tab:example_table}
\end{table}

The integration of the Directed Acyclic Graph Task Decomposition module with Retrieval-Augmented Generation technology resulted in performance improvements of 0.86\% and 1.3\% on MedQA and BioASQ tasks, respectively, relative to the isolated application of Retrieval-Augmented Generation. These performance enhancements are predominantly attributed to the structured decomposition of complex queries, facilitating more effective access to a broader spectrum of background information by the retrieval-augmented generation technology. This observation suggests that decomposing queries into hierarchical sets of sub-tasks optimizes the retrieval process, consequently improving the relevance and accuracy of the retrieved information.

Consequently, the synergistic interaction of these four modules not only improves overall performance on medical domain tasks but also effectively attenuates the adverse effects of irrelevant retrieval content, thus enhancing the efficacy of Retrieval-Augmented Generation technology in medical applications.

\subsection{Comparative Corpus Retrieval}
\noindent
The efficacy of Retrieval-Augmented Generation (RAG) systems being intrinsically linked to the quality of the retrieval corpus, this study investigates the influence of diverse retrieval corpora on the Bailicai model's performance across multiple medical benchmarks. Employing the Bailicai model, exclusively fine-tuned with Medical Knowledge Injection as a baseline, the study evaluated the effects of five distinct corpora. As evidenced in Table 8, the PubMed corpus demonstrated the highest average score of 71.58\% among the five corpora, with Wikipedia achieving a score of 70.87\%. These results align with anticipated outcomes, underscoring the inherent content limitations of each corpus. Specifically, PubMed, a platform primarily utilized for retrieving biomedical and life science literature, provides comprehensive and current content, encompassing cutting-edge and in-depth medical knowledge. Conversely, Wikipedia, an online encyclopedia covering a broad spectrum of topics from science to history and popular culture, offers substantial information volume but demonstrates limited depth in the medical domain. Relative to the baseline, the PubMed corpus exclusively exhibited performance improvements across four datasets, with a marginal decline of 0.18\% on the MMLU-Med dataset, thus substantiating the robustness of its corpus.

\begin{table*}[htb]
	\setlength{\belowcaptionskip}{0cm}
	\centering
	\caption{\textbf{Comparative Analysis of Bailicai's Performance Across Diverse Medical Corpora.}}
	\label{tab:methods}
	\begin{tabular}{@{}lccccccc@{}}
		\toprule
		\bfseries Corpus & \bfseries MedQA & \bfseries MedMCQA & \bfseries MMLU-Med & \bfseries PubMedQA & \bfseries BioASQ & \bfseries Average \\
		\midrule 
		+ Medical Knowledge Injection &65.67 & 55.82 & 75.85 & 77.40 & 78.80 & 70.71\\
		\midrule
		PubMed & \bfseries66.06 & \bfseries56.08 & 75.67& 78.40 & 81.72& \bfseries71.58\\
		Wikipedia &65.91&55.75&\bfseries76.40&77.00& 79.29& 70.87\\
		StatPearls &64.73&55.39&76.03&75.60&79.13& 70.18\\
		Textbooks&65.83&55.70&75.30&77.40&77.99& 70.44\\
		Merge &64.26&\bfseries56.08&76.03&\bfseries79.00&\bfseries81.88&71.45\\
		\bottomrule
	\end{tabular}
\end{table*}

\subsection{Comparative Document Retrieval Volume}
\noindent
The efficacy of retrieval-augmented generation (RAG) systems is substantially modulated by the quantity of documents retrieved. Fig. 3 elucidates a pronounced curve for the Bailicai system, delineating the correlation between the number of retrieved medical documents and accuracy across diverse medical benchmarks. The graph reveals that Bailicai maintains notable performance even with a minimal number of retrieved documents (e.g., one document), despite limited RAG-derived information, a phenomenon attributable to the medical knowledge integrated during the fine-tuning phase. As the number of retrieved documents increases, Bailicai's accuracy exhibits marginal enhancement and stabilization, consequent to the continuous augmentation of background information. This trend deviates from typical observations in RAG studies, where accuracy often fluctuates with increasing document retrieval, thus accentuating the robustness of the Bailicai system. Moreover, given the custom token length limit (2812) established during the fine-tuning phase, surpassing a specified threshold in the number of retrieved documents is hypothesized to induce a decline in accuracy.
\begin{figure}[ht!]
	\centering
	\includegraphics[width=250pt]{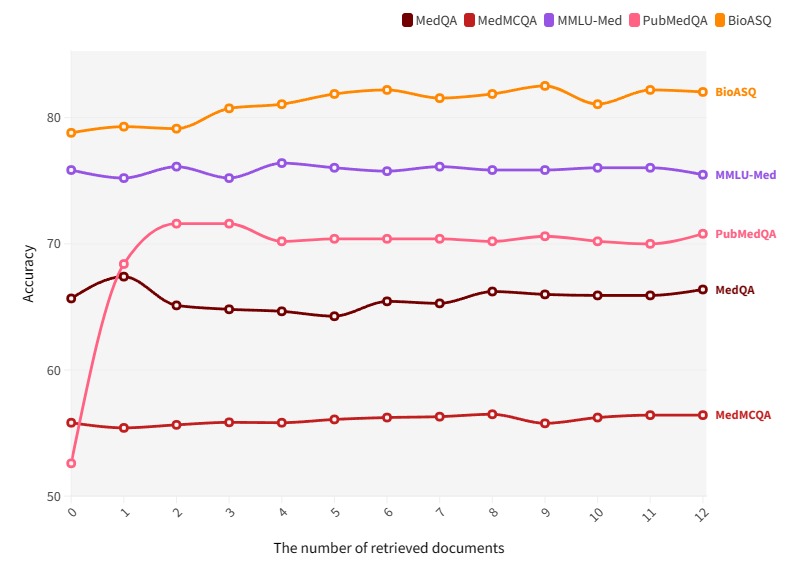} 
	\caption{Performance of Bailicai on Various Benchmarks with Different Retrieved Documents}
	\label{fig:myimage}
\end{figure}

\begin{table}[htb]
	\setlength{\belowcaptionskip}{0cm}
	\centering
	\caption{Statistics of 'Know' and 'Uknow' Categories and Average Number of Sub-questions Across Various Medical Benchmarks on Bailicai.}	
	\rowcolors{1}{gray!18}{white}
	\begin{tabular}{lccc} 
		\toprule
		Benchmark & Know& Unknow & Average Subproblems\\ 
		\midrule
		MedQA & 807 & 466 & 3.56\\
		MedMCQA & 3583 & 600 & 2.64\\
		MMLU-Med &956 &133&3.20\\
		PubMedQA & 169&331&1.78\\
		BioASQ & 217&401&2.81\\
		\bottomrule
	\end{tabular}
	\label{tab:example_table}
\end{table}

\subsection{Time Expense Analysis}
\noindent
In current research on Retrieval-Augmented Generation (RAG), retrieval is universally applied to all queries, neglecting the parameterized knowledge inherent in the model. While some studies have implemented a 'divide and conquer' approach, the processing of sub-queries remains deficient in targeted retrieval, resulting in inefficient resource allocation and increased computational time. To address this limitation, the Bailicai framework utilizes a model for Self-Knowledge Boundary Identification to ascertain whether an incoming query can be resolved without external knowledge, thus adaptively invoking RAG, analogous to a valve-switch mechanism. This methodology aims to improve computational efficiency and optimize resource allocation.

Table 9 illustrates the classification statistics for "unknown" and "known" queries, in addition to the mean number of sub-queries per benchmark test, as determined by the Self-Knowledge Boundary Identification model. By leveraging this model, the Bailicai framework can allocate resources more efficiently to complex and challenging queries, thereby reducing the frequency of retrieval system calls and improving performance when integrated with other frameworks. 

Moreover, the MedQA benchmark, which encompasses extensive patient inquiry data, exhibited the highest mean number of sub-queries at 3.56. This benchmark illustrates the model's proficiency in distinguishing and decomposing query complexity. The decomposition of patient inquiries into granular sub-tasks enables the collection of critical clinical indicators necessary for diagnosis, thus improving the efficacy of problem resolution. These results are further substantiated in Table 6.

\section{Conclusions}
\noindent
In this paper, we present the Bailicai Framework, a retrieval-augmented generation framework specifically tailored for the medical field to improve medical outcomes. The Bailicai Framework is engineered to autonomously perform retrieval tasks and decompose complex queries into a structured hierarchy of sub-problems. Furthermore, this framework incorporates content evaluation mechanisms to mitigate hallucination phenomena prevalent in large language models utilized in medical applications, while concurrently reducing noise stemming from irrelevant or spurious documents typically encountered in conventional retrieval-augmented generation methodologies. Empirical evaluations across diverse medical benchmarks corroborate the superior performance of the Bailicai Framework. Ablation studies further substantiate the efficacy of the framework's constituent components.

\end{document}